\documentclass[runningheads]{llncs}
\usepackage{graphicx}
\usepackage{hyperref}
\usepackage{xcolor}
\usepackage{xspace}
\usepackage{newtxtext}

\usepackage{paralist}
\usepackage{enumitem}
\usepackage{booktabs}

\newcommand{\aim}[1]{$\mathbf{AIM_{#1}}$}

\newcommand{\cq}[1]{$\mathbf{CQ_{#1}}$}
\newcommand{\rdf}[1]{\textit{#1}\xspace}
\newcommand{\modality}{\rdf{Modality}}

\newcommand{\modaldescriptor}{\rdf{ModalDescriptor}}
\newcommand{\mmentity}{\rdf{Multi-ModalEntity}}
\newcommand{\modalityimage}{\rdf{ImageModality}}
\newcommand{\modalityvideo}{\rdf{VideoModality}}
\newcommand{\modalitytext}{\rdf{TextModality}}

\newcommand{\hasmodality}{\rdf{hasModalitySpecification}}
\newcommand{\hasmodaldescriptor}{\rdf{hasModalDescriptor}}

\newcommand{\resourceURL}{\rdf{resourceURL}}
\newcommand{\hasPathInArtifact}{\rdf{hasPathInArtifact}}

\newcommand{\imgobj}{\rdf{ImageObject}}
\newcommand{\imgdata}{\rdf{ImageData}}

\begin{document}
\title{A Pattern to Align Them All: Integrating Different Modalities to Define Multi-Modal Entities}

%
\titlerunning{A Pattern to Align Them All}
%

\author{Gianluca Apriceno\inst{1} \orcidID{0000-0003-4603-6888} \and
Valentina Tamma \inst{2}\orcidID{0000-0002-1320-610X}
\and
Tania Bailoni\inst{1}\orcidID{0000-0003-4360-566X}
\and
Jacopo de Berardinis\inst{2}\orcidID{0000-0001-6770-1969}
\and
Mauro Dragoni\inst{1}\orcidID{0000-0003-0380-6571}
}
\authorrunning{Apriceno et al.}
%
\institute{Fondazione Bruno Kessler, Trento, Italy\\ 
\email{\{apriceno, tbailoni, dragoni\}@fbk.eu}\\
\and University of Liverpool, Liverpool, United Kingdom\\
\email{\{V.Tamma, jacodb\}@liverpool.ac.uk}
}
\maketitle              

\begin{abstract}
The ability to reason with and integrate different sensory inputs is the foundation underpinning human intelligence and it is the reason for the growing interest in modelling multi-modal information within Knowledge Graphs.

Multi-Modal Knowledge Graphs extend traditional Knowledge Graphs by associating an entity with its possible modal representations, including text, images, audio, and videos, all of which are used to convey the semantics of the entity. Despite the increasing attention that Multi-Modal Knowledge Graphs have received, there is a lack of consensus about the definitions and modelling of modalities, whose definition is often determined by application domains. 
In this paper, we propose a novel ontology design pattern that captures the separation of concerns between an entity (and the information it conveys), whose semantics can have different manifestations across different media, and its realisation in terms of a physical information entity. By introducing this abstract model, we aim to facilitate the harmonisation and integration of different existing multi-modal ontologies which is crucial for many intelligent applications across different domains spanning from medicine to digital humanities.

\\
\textbf{GitHub:} \href{https://github.com/IDA-FBK/MultiModalPattern}{https://github.com/IDA-FBK/MultiModalPattern}
\end{abstract}

\section{Introduction}
\label{sec:01-introduction}
There is an increasing demand for modelling \textit{multi-modal} aspects of knowledge to break through the bottleneck of real-world applications~\cite{Shi2019,Niu2021} or, more recently, to be exploited either for manipulating or customising Large Foundational Models~\cite{Cheng2023}. 
Indeed, as stated by Zhu and colleagues \cite{Zhu2024}, the ``multi-modalization'' of knowledge is a key step towards achieving human-level machine intelligence.

Knowledge Graphs (KGs) have had a fast growth in the last decade and the benefits deriving from their adoption have been perceived in a wide range of real-world applications, including text understanding, recommendation systems, and natural language question answering. 
The landscape of KGs includes resources covering common sense knowledge~\cite{Matuszek2006,Liu2004}, lexical knowledge~\cite{Miller1995,Navigli2010}, and encyclopedic knowledge~\cite{Auer2007,Suchanek2007,Vrandecic2014} to name the most salient ones. 
However, most of the existing KGs are mainly represented in symbolic form, which may weaken the capability of machines to describe and understand the real world. 
The intrinsic understanding of the concept ``pizza'' in humans is inextricably linked to the sensory experiences involved in its preparation and consumption.
This observation underscores the importance of grounding symbols like "pizza" to their real-world referents~\cite{Harnad1990,Steels2008}.
Symbol grounding, which involves establishing connections between symbols and corresponding sensory data such as images, sounds, and videos, is crucial for artificial agents to approximate human-like behaviour ~\cite{Harnad1990}.
By mapping symbols to their corresponding referents in the physical world, these agents can potentially develop a deeper understanding of concepts, thereby enhancing their ability to interact with and reason about the world.

A way to address the need for such grounding is the construction of Multi-Modal Knowledge Graphs (MMKGs)~\cite{Peng2023}, where each knowledge symbol can be associated with its corresponding data items in modalities other than text, such as image, sound, or video, that could embody the knowledge.

Different approaches have been proposed in the literature to engineer MMKGs.
However, there is a lack of consensus on what a modality is, with the definition of modality strictly connected to its content. General purpose vocabularies (e.g. Schema.org~\footnote{\url{https://schema.org}} and foundational ontologies, e.g. DOLCE Ultra Lite Plus~\cite{presutti2016dolce+} do not make any form of commitment regarding whether a modality and its content should be distinguishable and the meaning each of these convey. In other approaches, such as \cite{Bachvarova2007}, the meaning carried by a modality is determined by its content (i.e. the particular information it represents), by its content-independent characteristics (i.e. its nature), and the relations existing between these two main aspects.

Within this context, the purpose of the paper is two-fold.
Firstly, we provide a conceptual and systematic analysis of multi-modality by starting from the requirements motivating the need to harmonise the notion of multi-modality from the semantic perspective.
Secondly, we propose a conceptual pattern satisfying the requirements identified, which in turn
provide an abstract (upper) ontology on multi-modality that aligns and connects existing (multi-modal) ontologies, and where the definition of modality (referred in the paper as \textit{nature}), is explicitly described together with its content (i.e., the digital resource). Therefore,  we aim to propose a conceptual analysis and a possible abstract schema to foster the research activities on MMKG. Our intent is not to propose a new resource since the upper ontology described in the next sections has not been used yet by the community to bridge different modelling choices around MMKGs.

The remainder of the paper is organised as follows.
In Section~\ref{sec:02-background} we review relevant work on ontologies and knowledge engineering methodologies for building MMKGs; Section~\ref{sec:03-motivation} explains the motivation of our work; Section~\ref{sec:04-pattern} presents the pattern while Section~\ref{sec:05-application} validates our proposition by demonstrating its applicability to current MMKG resources as well as the alignment the existing ontologies with it.
Finally, in Section~\ref{sec:06-conclusion} we draw concluding remarks and outline directions for future works.

\section{Background}
\label{sec:02-background}

Autonomous intelligent behaviour relies on the ability to perceive multi-modal information from the environment, and reason with this information, possibly aided by some form of background or common-sense knowledge to support decision-making and distil new knowledge~\cite{lecun2022path}.
\textit{Multi-modality} has been defined as ``the coexistence of more than one semiotic mode within a given context''~\cite{gibbons2012multimodality}, something that living beings experience in everyday life through sight, sound, and movement.
A similar coexistence of modalities has also been adopted by many robotics and autonomous systems, whose decision-making is based on data generated by a plethora of different sensor types, e.g. vision, proximity, and tactile sensors~\cite{calbimonte_et_al:TGDK.1.1.13}. 
This has resulted in the proliferation of multi-modal information in the form of images, sounds, or sensor-based location information, to name a few~\cite{Peng2023}, typically represented using different data formats. 

Intelligent decision-making~\cite{Zhu2024}, therefore, relies on processing multi-modal knowledge such that symbols are grounded in their corresponding modalities and mapped to their referents together with their meaning.
Indeed, without the ability to ground symbols to the modality they refer to, and to map them to their corresponding referents with meaning (therefore associating specific data elements to their real-world counterparts in a manner that preserves their semantic relevance or meaning) a machine has only a limited understanding of its environment. 


KGs are gaining prominence as a formalism to model and process multi-modal knowledge, and therefore provide an explicit representation of the association between data elements and their semantic referents.
MMKGs extend traditional KGs by integrating information from diverse modalities, thus representing different modal aspects that contribute to the definition of the same entity.
In MMKGs, some of the symbols used to denote nodes and edges together with their corresponding data items, are associated with diverse modalities such as text, image, sound, or video, that could embody the knowledge~\cite{Zhu2024}. 
The inclusion of different modalities provides a more comprehensive, and contextually aware view of entities modelling a domain, therefore better capturing the nuances and complexity of real-world data, and providing a richer understanding of the relationships connecting them~\cite{Sheth2019,Chen2024KnowledgeGM}. 
Indeed, these modalities rarely exist in isolation but are often intertwined with structured semantic information, in an application-dependent fashion. 
Numerous efforts in the literature leverage this interplay across a range of tasks, such as image retrieval ~\cite{Wang2008,Khalid2011,Poslad2014}, Image Captioning~\cite{Lu2018EntityAware,Huang2020ImgCaptioning,Zhang2020radiology,Zhao2024boosting} and Visual Storytelling \cite{Hsu2021PlotAndRework,Xu2021ImagineReasonWrite,Li2023KnowledgeEnrichedAttention}



Despite this growing interest in modelling multi-modal data, the term \textit{modality} does not have a widely accepted definition. 
In particular, differences in semantics arise when multi-modal information is encoded in KGs. 
In most MMKGs, the concept of modality remains tightly intertwined with the corresponding (digital) resource being modelled. A notable example is \texttt{schema.org}, where the nature of both \texttt{MediaObject} and its more specific subclasses, such as \texttt{ImageObject} and \texttt{AudioObject}, is closely linked to its content, as indicated by the \texttt{contentURL} property.

In the context of this work, we argue the separation of concerns between the resource being modelled, and \textit{how} this is modelled, which can manifest itself through different formats: images, text, etc, each contributing some of its meaning to a resource. We, therefore, consider a resource as a type of \textit{Information Entity} (IE)~\cite{SanfilippoEmilioM.2021Ofie}, i.e. the \textit{content} shared by different copies of artefacts such as narrative fictions, musical scores, personal records, images or engineering specifications (e.g design models or process plans). The concerns underlying the modelling of IEs have been addressed in different disciplines from philosophy to linguistics, to applied ontology~\footnote{We refer to \cite{SanfilippoEmilioM.2021Ofie} for a thorough presentation of the state of the art of ontologies for IEs.}
Information entities (IEs) are social objects created and used for communication, reasoning, and the specification of new entities. These entities exist within a social context and serve various functions in the exchange and processing of information.

An \textit{Information Object} (IO) is a specific type of non-physical IE.
It is a container of information that can exist independently of its physical manifestations.
Examples of IOs include the content of a musical composition, the plot of a novel, or the instructions of a recipe.
While the IO itself is intangible, it can be realised through various physical means. An \textit{Information Realisation} (IR) is the physical manifestation or embodiment of an information object.
It is the tangible form through which an IO is expressed, communicated, or experienced.
In the case of a musical composition, the IR could be a live performance, a recording, or sheet music \cite{de2023music}; for a novel, the IR might be a printed book, an ebook, or an audiobook.
This distinction between IOs and IRs is crucial as it allows us to separate the informational content from its material representation.
From this perspective, \texttt{schema.org}'s \texttt{MediaObject} can be broadly considered as an information entity \cite{infoRealizationODP}, however, its definition blends/encapsulates elements of both IO (stemming from the subclassing of \texttt{CreativeWork}) and IR (encoding format, bitrate, player type, etc.).
While this ambivalent characterisation provides flexibility for direct use, it may also introduce inconsistencies (e.g. a MediaObject may be associated with a CreativeWork which is different to the one it encapsulates).
In addition, this violates the Information Realisation ODP \cite{infoRealizationODP} -- as information objects and realisations are defined as disjoint classes.
Similarly, in the Ontology for Multimodal Knowledge Graphs for Data Spaces~\cite{Usmani2023}, the \textit{mmkg:Source} concept encompasses both the scene file and the type of modality it encapsulates.

The separation between Information Object and its realisation is also adopted in ~\cite{Bachvarova2007}, who proposed a distinction between the specification of a modality (i.e., the properties contributing to its definition) and the data content.
Their definition focused on ``traditional'' modalities, whereas we aim for a comprehensive ODP that provides modelling flexibility while accommodating composite and new modalities (not anticipated at the time of writing), including unconventional and emerging ones (see Section~\ref{sec:03-motivation}).


\section{Motivation}
\label{sec:03-motivation}

Despite the increased popularity of MMKGs, an analysis of the literature  identified three main problems: 
\begin{enumerate}
\item the term \textit{modality} does not have a well-accepted definition and consequent ontological formalisation (except for the work in \cite{Bachvarova2007}). The use in the literature can encompass both typical examples of modalities, e.g. images, sounds, or less typical ones, e.g. linguistic features of co-occurring terms~\cite{Kress2010,Agnieszka2015,Gao2019,Lu2020,Zhang2020}; \item typical MMKGs tend to model homogeneous knowledge, i.e. either images, movies, or sound, which is inherently siloed~\cite{Zhu2024}. Some recent state-of-the-art studies, e.g.~\cite{10.1145/3656579}, focus primarily on textual and image information individually and do not consider the possible interaction between modalities. This homogeneity affects also representational learning~\cite{NIPS2013_1cecc7a7}, where many multi-modal KG embedding approaches map all multi-modal information onto one high-dimensional embedding~\cite{Pezeshkpoure-et-al2018}; and
\item the scope and the type of the knowledge modelled typically depends on the task for which it is designed, and therefore is not easily reused in or aligned to other applications~\cite{Parcalabescu2023}.
\end{enumerate}


To overcome these limitations, we propose a novel \textit{Multi-modality pattern}, that is general in scope and supports the coexistence of different modalities.
Ontology design patterns (ODPs) can be seen as a lightweight module or fragment of a foundational ontology, therefore providing an abstraction of sound modelling choices that can be shared and reused across ontologies or KG schemata~\cite{keet2018introduction}. 
The proposed pattern establishes the ontological commitment to the notion of modality and thus separates the manifestation of the modality from its semantics.

The design of this pattern~\cite{hammar2017content,blomqvist2021advances} is based on a set of requirements that MMKGs should satisfy, and that were identified by considering two main activity types in the KG engineering life-cycle, MMKG creation and MMKG alignment. 

\textit{MMKG creation} refers to the instantiation of a MMKG.
In an MMKG, the schema for a particular task or application is a specialisation of the proposed Multi-Modality pattern.
The pattern provides also the blueprint for defining how to instantiate the schema, in the creation and evolution of a KG of interconnected multi-modal entities, where each entity may be described by more than one modality.
We explore this further in Subsection~\ref{subsec:5.1-FusKG}. 
By \textit{MMKG alignment} we refer to those cases in which it is necessary to align and integrate several existing MMKGs that were developed for some downstream application to represent a more exhaustive view of some multi-modal entity, and to support interoperability between multi-modal representations. For example, we could align the entity representing ``Hey Jude'' in KGRec-sound, a dataset of songs with song's textual descriptions, tags and user’s listening habits extracted from Songfacts.com and Last.fm, respectively~\cite{10.1145/2926718} with the corresponding DBpedia~\cite{Auer2007} entity that links the images depicting the song, e.g. the picture of Julian Lennon, for whom the song was written.  
We explore this further in Subsection~\ref{subsec:5.2-Alignments}. 

While these activity types are not meant to be exhaustive they allow us to show the applicability, generality, and robustness of the pattern for different situations.

We use these two activities to identify requirements for the design of the Multi-modal pattern we propose: 

\begin{enumerate}
\item the pattern should support the integration of different modalities that model \textit{holistically} the same entity;
\item the pattern should promote semantic agreement on the definition of \texttt{modality};
\item the pattern should represent modalities at the highest level of abstraction, therefore ensuring a separation of concerns between the specific downstream applications supported and the content of the pattern;
\item the pattern should be domain agnostic, task-independent and easy to use;
\item the pattern should promote modular development and be extensible, therefore allowing users to define new modalities and express relationships between them.
\end{enumerate}

\section{The Multi-Modal Ontology Design Pattern}
\label{sec:04-pattern}

The \textit{Multi-modal} ODP's goal is to represent information entities with their associated modalities, of diverse nature, and the corresponding relationships between them (\aim{1}) while providing flexibility and extensibility to seamlessly incorporate new modalities and adapt to changes in existing ones (\aim{2}). These two aims determine the requirements underlying the pattern's design:
\aim{1} identifies the functional requirements that the pattern should satisfy and that we represent in the four competency questions (\cq{s}) outlined in Table~\ref{tab:competency-questions}.
%
%
\aim{2} relates to the set of (non-functional) requirements identified in Section~\ref{sec:03-motivation}.
Overall, our proposition extends and combines the expressivity of the Information Realization ODP \cite{infoRealizationODP} with the flexibility of the modality separation approach advocated by \cite{Bachvarova2007}.

\begin{table}[h!]
\centering
\begin{tabular}{@{}p{1cm}l@{}}
\toprule
\textbf{ID} & \textbf{Competency question} \\ \midrule
CQ1 & What are the modalities that interact in the definition of a data resource? \\
CQ2 & If applicable, which other modalities do a modality subsume? \\
CQ3 & What are the relation(s) between a resource and the modalities associated with it? \\
CQ4 & What is the format associated with a resource? \\
\bottomrule
\end{tabular}
\caption{Competency questions driving the design of the Multi-modal ODP}
\label{tab:competency-questions}
\end{table}

\begin{figure}[h]
\centering
    \includegraphics[scale=0.33]{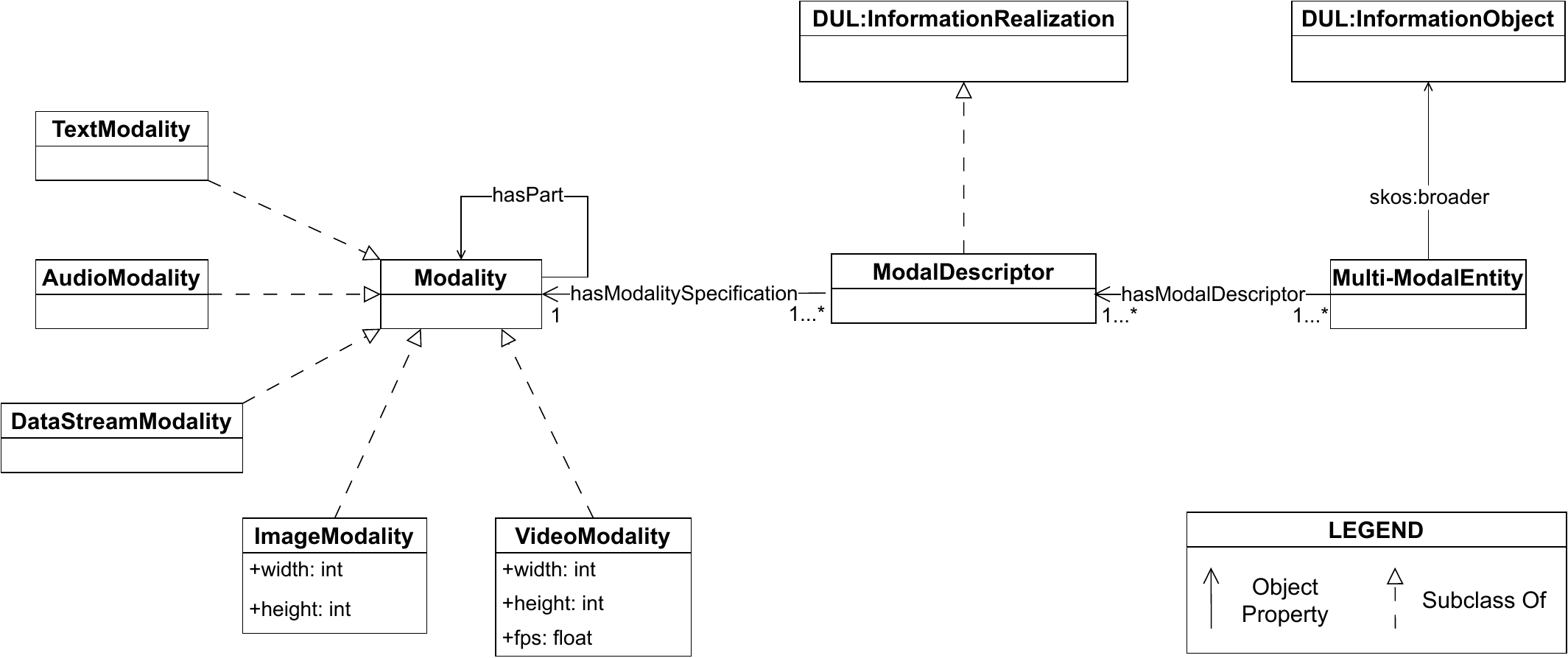}
    \caption{Conceptual model of the Multi-modal ODP.}
    \label{fig:conceptualization_mm}
\end{figure}

Figure~\ref{fig:conceptualization_mm} illustrates the proposed ODP, showing the interplay between the 3 main classes: the multi-modal entity (\mmentity), the modality-specific realisations associated with the former (\modaldescriptor), and the modality specification of a modal descriptor (\modality).
For instance, in the culinary context, a recipe (an instance of \mmentity) can be linked to different types of multi-modal resources, such as the text listing the steps to prepare the recipe\footnote{With ``text'' we refer to the raw textual (content) description of a multi-modal entity}, the video showing its preparation steps, or the images of the finished dish.
Each of these realisations is associated with a specific modality.

\mmentity is defined as any entity or concept that can be associated with a modal representation 
that contributes to its description.
It has a broad scope by design, to provide the flexibility required for ontology instantiation, e.g. as it is represented in Wikidata and \texttt{schema.org} via open domain relationships to multi-modal objects (see for example \href{http://www.overleaf.com}{P18} and \url{https://schema.org/image}).
In most cases, a \mmentity is an Information Object (IO, c.f. Section~\ref{sec:02-background}), such as a recipe, or the content of a musical composition, but it can also be a living entity (e.g. a person), or an abstract state (e.g. a mood or emotion).
While, in principle, anything can be associated with a modal realisation (possibly originating from another IO), the specific scoping of a \mmentity depends on the target application and context.

The \modaldescriptor is a data object that is uniquely assigned to a modality.
It is a specific type of Information Realization (IR), as it refers to information entities that are digitally represented and can be stored as files.
For example, possible IRs of a musical entity also include performances and improvisations, in addition to audio recordings and scores, as described in Section~\ref{sec:02-background}. However, only the latter two are valid examples of a \modaldescriptor.
%
The \modaldescriptor can be seen as the reification of the relationships among \mmentity (i.e., the entity with one or more modal resources), the associated resources, and their modalities.
These relationships are defined through the \hasmodaldescriptor object property having \mmentity as domain and \modaldescriptor as range; and the \hasmodality object property having \modaldescriptor as domain and \modality as range.
Depending on the usage of the pattern, one may define a \resourceURL datatype property as a pointer to the multi-modal resource available on the Web (e.g., the URL of an image), or simply align the URI of a \modaldescriptor with other entities.
Similarly, a \hasPathInArtifact datatype property may be used when the multi-modal resource is contained within host packages, e.g., a Zip archive. 
In such cases, a \resourceURL would provide the URL of the Zip archive, while the \hasPathInArtifact would contain the specific path within the multi-modal resource within the Zip archive.

The \modality class is the most abstract representation of a modality.
As an abstract class, \modality itself cannot be directly instantiated.
Examples of specific modalities in the ODP include \modalitytext, \modalityimage, and \modalityvideo, which are all defined as subclasses of \modality.
Specific concepts for each modality type enable the modelling of individual instances sharing common properties.
For instance, an individual \rdf{Image\_jpeg\_430x320}, instantiating \modalityimage, possesses characteristics such as width (430 pixels), height (320 pixels), and jpeg format (annotation \textit{dc:format}~\footnote{\href{https://www.dublincore.org/specifications/dublin-core/dcmi-terms/terms/format/}{https://www.dublincore.org/specifications/dublin-core/dcmi-terms/terms/format/}}).
This approach facilitates the definition of relationships between modalities, allowing reuse and extension of elements defined for other modalities.
Additionally, a reflexive property \texttt{hasPart} is introduced for the \modality concept.
This parthood relationship, of type containment~\cite{Keet2008RepresentingAR}, allows the modelling of complex modalities in terms of atomic ones, where one modality embeds others.
For example, an online recipe note might initially consist solely of \modalitytext, but the addition of an image establishes a complex modality where the \texttt{hasPart} relationship expresses the containment of \modalitytext and \modalityimage.

Overall, our modelling choices achieve \aim{2}, as any changes to existing modalities can be accommodated by extending the \modality subclasses.
This also includes shared and general changes that affect \modality, as well as more specific modifications involving its subclasses.
Finally, the pattern also allows for the introduction of new modalities by adding corresponding subclasses to \modality, thereby supporting future extensibility.

\section{Applications and Alignments}
\label{sec:05-application}
This section demonstrates the application of the ODP pattern within an existing KG and illustrates how current multi-modal ontologies are aligned through its utilisation.

\subsection{Multi-Modal Pattern Application: FuS-KG}
\label{subsec:5.1-FusKG}

The Functional Status Knowledge Graph (FuS-KG) is a multi-modal knowledge graph that provides a complete and structured way to represent all the information about the function status information (FSI) of a person. 
This is done with the aim of building and deploying more effective coaching solutions that, by exploiting the FuS-KG, can support fragile individuals during their daily lives in achieving a healthy lifestyle.
Currently, FuS-KG comprises ten distinct modules: \textit{Core}, \textit{Food}, \textit{Recipes},  \textit{Diseases}, \textit{Activity}, \textit{Barrier} \textit{Temporal}, \textit{User}, \textit{Guideline}, and \textit{Multi-modal}. 
Notably, the multi-modal module, which contains multi-modal recipe resources (e.g., images and videos of recipes), has been implemented in accordance with the ODP of Section~\ref{sec:04-pattern}. We present two illustrative examples that show how the ODP is used within the module.

\subsubsection{Image modality:}
The recipe shown in Figure~\ref{fig:mm_recipe1m} refers to the individual named Recipe1m-60 (namely, Crock Pot Caramelized Onion). 
This recipe is associated with two individuals of type \modaldescriptor, i.e., Recipe1m-60-image\_jpeg480x320-1 and Recipe1m-60-image\_jpeg 550x826-1 18, each one pointing to different \modalityimage modalities. 
It is important to note that the naming convention used for \modaldescriptor
instances includes the recipe name, the modality metadata (e.g., image, video, and format), and a progressive number. 
This number aims to disambiguate multiple instances associated with the same modality within a recipe. 
Indeed, a recipe may have multiple images that share image modalities.
For example, Recipe1m-60 may have multiple images of the same format and size, leading to sequentially numbered modal descriptors (e.g., Recipe1m-60-image jpeg 480x320-2 sharing the image-jpeg 480x320 modality. 
As can be seen, each \modaldescriptor instance is provided with the URL data properties pointing to the corresponding (image) resource.

\subsubsection{Video modality:}
\begin{figure}[!t]
\centering
    \includegraphics[scale=0.42]{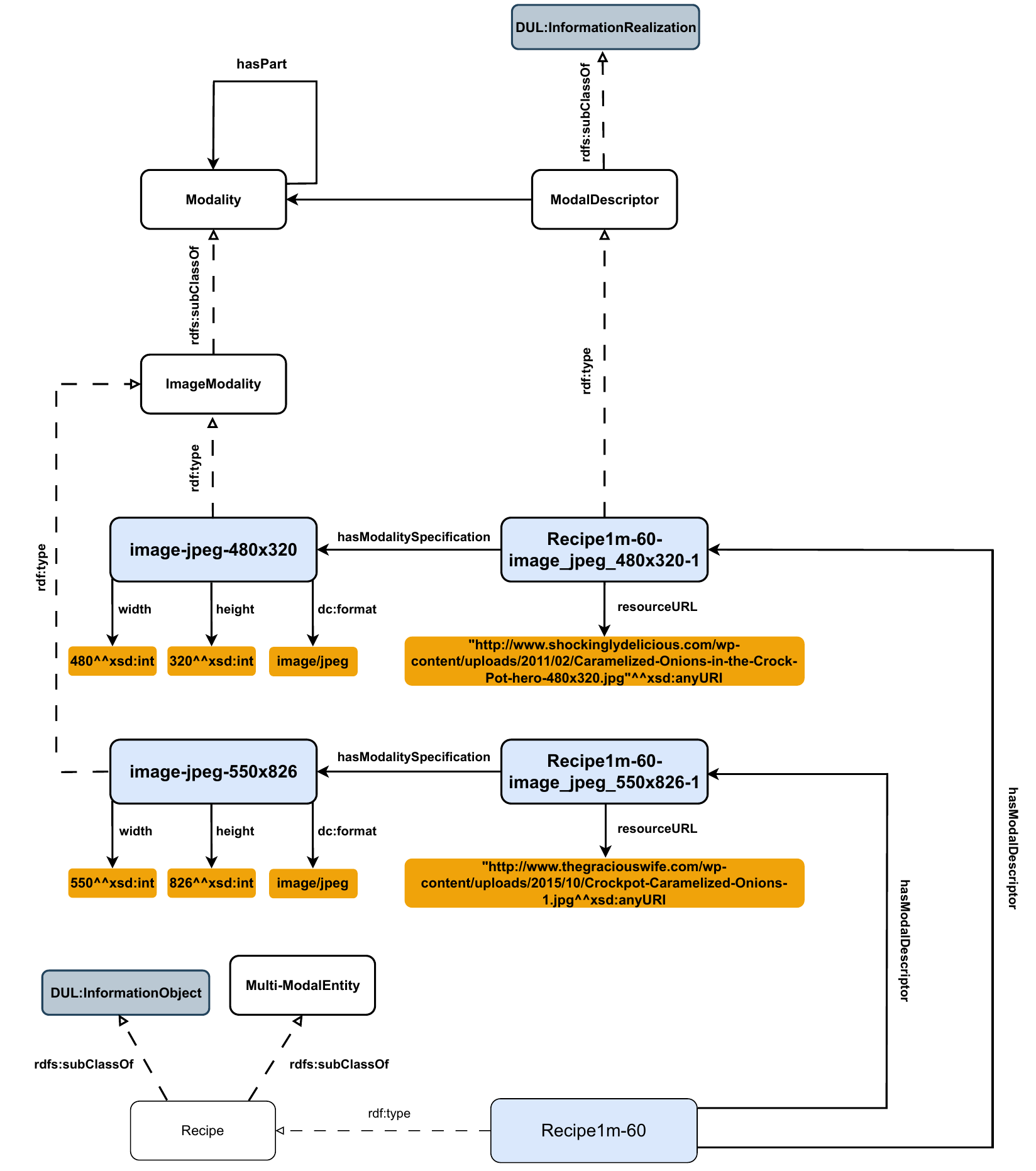}
    \caption{Application of the ODP in FuS-KG (Image modality).}
    \label{fig:mm_recipe1m}
\end{figure}

\begin{figure}[!ht]
\centering
    \includegraphics[scale=0.45]{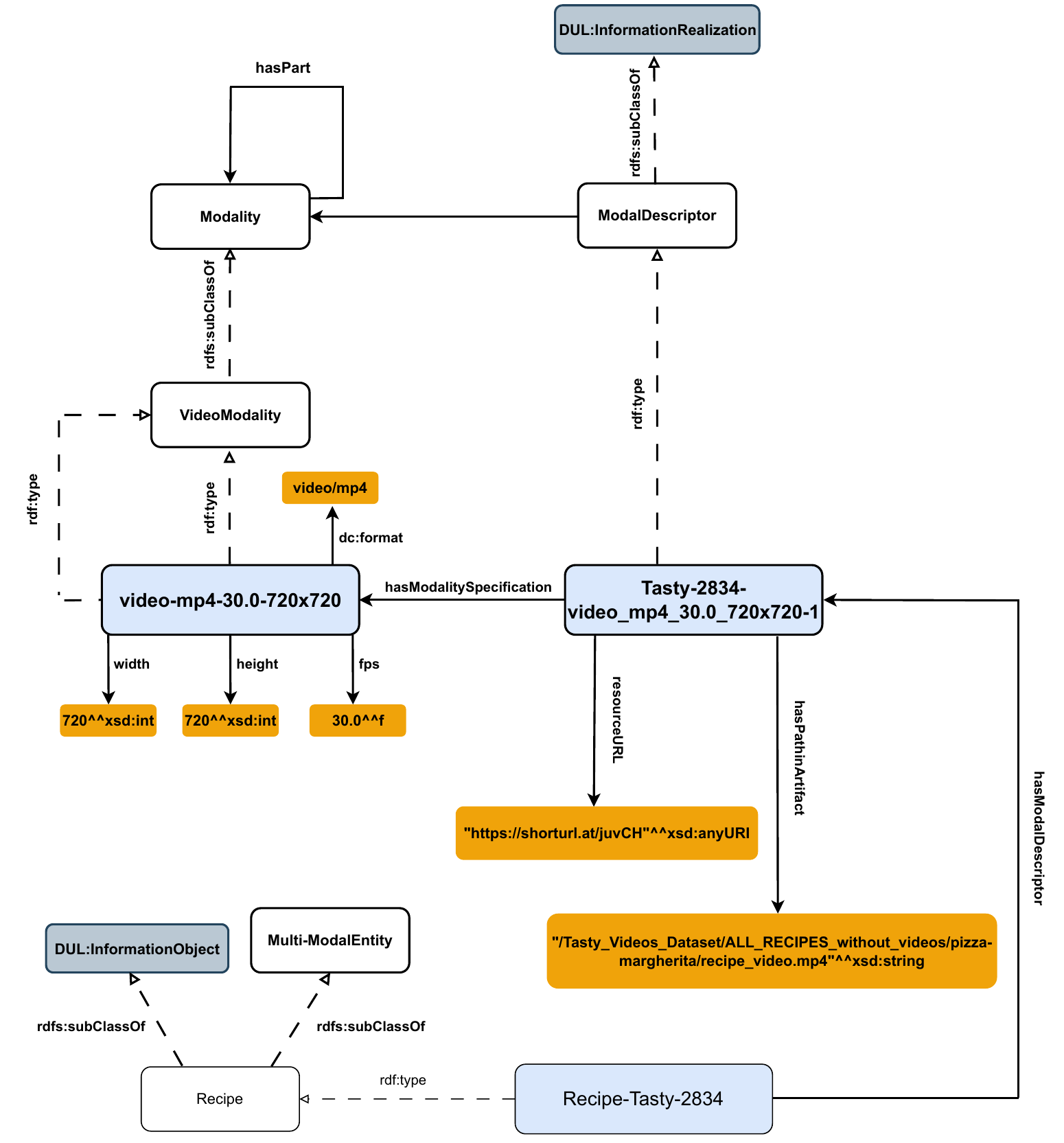}
    \caption{Application of the ODP in FuS-KG (Video modality).}
    \label{fig:mm_tasty}
\end{figure}

The recipe shown in Figure~\ref{fig:mm_tasty} refers to the individual named Tasty-2834 (namely, Pizza Margherita). 
This recipe is associated with one individual of type \modaldescriptor (i.e., Tasty-2834-video\_mp4\_30.0\_720x720-1) pointing to a \modalityvideo. 
Differently from Figure~\ref{fig:mm_recipe1m}, the naming convention for \modaldescriptor instances, in this case, also contains the frames-per-second (fps) that is used as a data property to represent the frames per second of a specific video instance. 
Furthermore, this recipe instance, differently from the one shown in Figure~\ref{fig:mm_recipe1m}, represents an example of an instance where the resource (i.e, the video of the recipe), is contained in a specific path inside an (external) zip archive as denoted by the data properties, \hasPathInArtifact, and \resourceURL, respectively.



\subsection{Multi-Modal Pattern Alignment: Aligning Existing Ontologies}
\label{subsec:5.2-Alignments}
In this subsection, we show how existing ontologies covering multi-modality, can be aligned with our ODP. 
We select and show the alignment for the following ontologies:
\begin{itemize}
    \item The Multimodal Description of Social Concepts Ontology (MUSCO) \cite{Pandiani2021}
    \item The Ontology for Multimodal Knowledge Graphs for Data Spaces \cite{Usmani2023}
    \item The Polifonia Ontology Network (PON) \cite{Berardinis2023} 
\end{itemize}


\subsubsection{MUSCO:}
\label{subsubsec:MUSCO}
Automatic detection of social concepts evoked by images presents a significant challenge. 
This is primarily due to the complexity of social concepts, which lack unique physical features and rely on more abstract characteristics compared to concrete concepts. 
The Multimodal Description of Social Concept (MUSCO) ontology has been proposed to investigate, model, and experiment with how and why social concepts (e.g., peace and violence) are represented and detected by both humans and machines in images. 
To achieve this, the ontology integrates multisensory data and is grounded in cognitive theories about social concepts, mapping them to multimodal frames. 

\begin{figure}[!t]
\centering
    \includegraphics[scale=0.44]{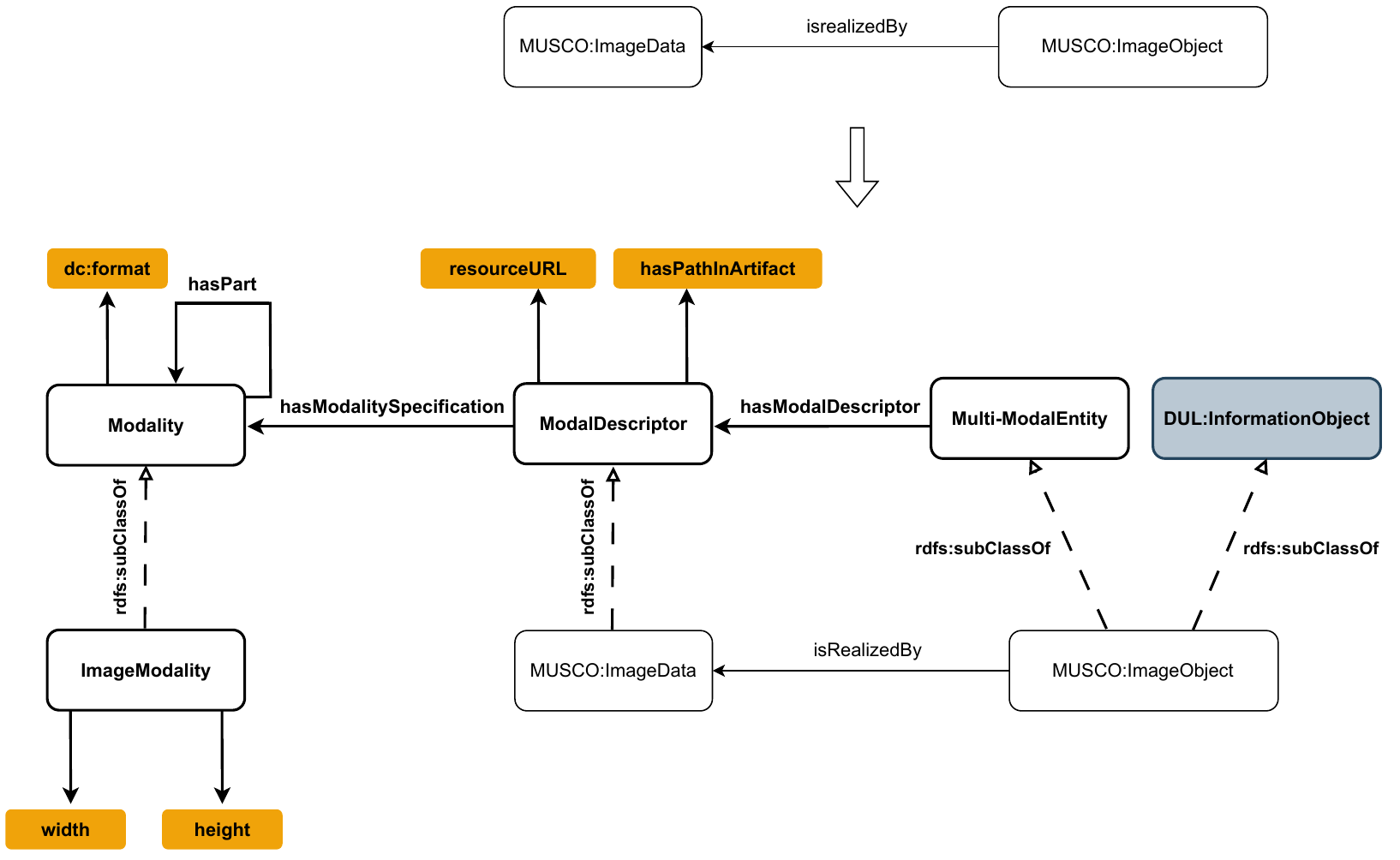}
    \caption{Alignment of the ODP with MUSCO.}
    \label{fig:alignment_with_musco}
\end{figure}

The upper part of Figure~\ref{fig:alignment_with_musco} illustrates the relationship between the \imgobj and the \imgdata concepts within MUSCO ontology.
The definition of \imgobj designates it as a type of information object with a location in a pixel box and indicates its use in evoking a social concept (i.e., \textit{SCMultiModalFrame} within MUSCO). In contrast, ImageData is treated as one of its possible (information) realisations. 

The lower part of Figure~\ref{fig:alignment_with_musco} shows the alignment of these concepts within the ODP shown in Figure~\ref{fig:conceptualization_mm}. 
The \imgdata has been aligned by specifying it as a subclass of the \modaldescriptor class. 
Additionally, the content of the \imgdata has been distinguished from its modality through the use of the \modalityimage modality. 
The \imgobj encompasses the role of the \mmentity, presented in Figure~\ref{fig:conceptualization_mm}. 
Consequently, this class inherits the \hasmodaldescriptor property, which connects the \imgobj with its associated image resource(s).


\subsubsection{Ontology for Multimodal Knowledge Graphs for Data Spaces:}
\label{subsubsec:DataSpaces}
Data spaces~\cite{Haley2006dataspaces} offer flexible and scalable solutions for integrating heterogeneous data from multiple, multi-modal sources. 
By combining structured and unstructured data, data spaces enable detailed insights into complex patterns hidden within the data.
Moreover, integrating external background knowledge can enhance the reasoning process by inferring new facts not immediately deducible from the data. 
In this context, MMKGs have the potential to enrich data spaces by providing a unified framework for reasoning over structured multi-modal data from diverse sources. 
To address this, \cite{Usmani2023} proposes a foundational multi-modal ontology to represent, query, and analyse multi-modal data in data spaces.

\begin{figure}[!t]
\centering
    \includegraphics[scale=0.44]{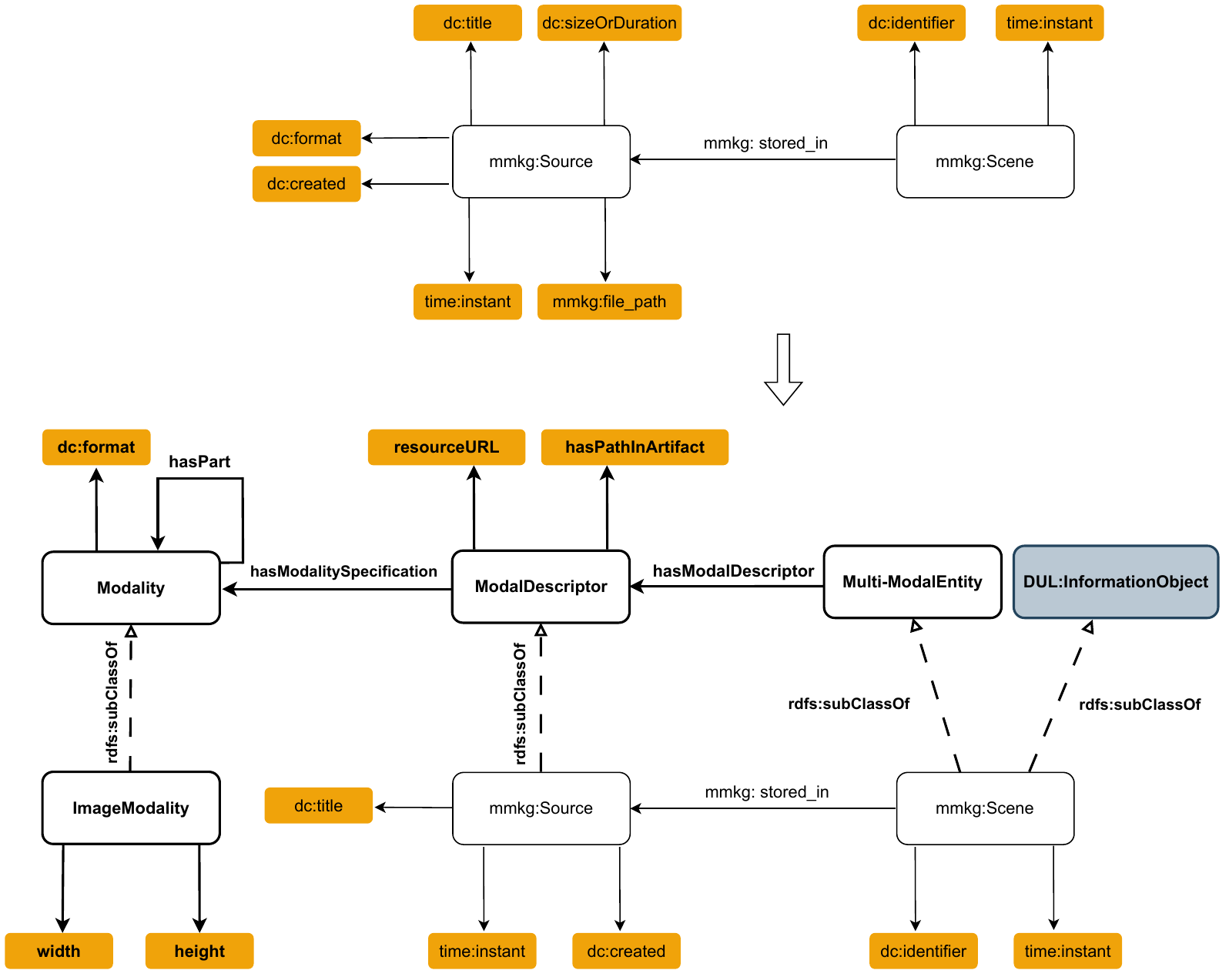}
    \caption{Alignment of the ODP with Ontology for Multimodal Knowledge Graphs for Data Spaces.}
    \label{fig:alignment_with_usmani}
\end{figure}

The upper part of Figure~\ref{fig:alignment_with_usmani} illustrates an excerpt of the ontology, depicting the relationship between the primary concepts involving multi-modality: \textit{mmkg:Scene} and \textit{mmkg:Source}. 
The \textit{Scene} class represents a scene captured by a camera, characterised by two data properties: \textit{dc:identifier} and \textit{time:instant}, denoting the timestamp of the scene captured. 
The \textit{Source} class represents unstructured data files received from the camera, such as images. 
This class contains properties describing a source, including \textit{dc:sizeOrDuration}, \textit{dc:title}, \textit{dc:format}, \textit{dc:created} (indicating the file creation time), \textit{time:instant}, and \textit{mmkg:filePath}, to denote the file's location. 
The \textit{Source} class aligns closely with \modaldescriptor, with the distinction that the nature of the \textit{Source} is intrinsically linked to its content.

The lower part of Figure~\ref{fig:alignment_with_usmani} shows the alignment of \textit{Source} and \textit{Scene} concepts within the ODP. 
As can be seen from the figure, the \textit{Source} class is specified as a subclass of \modaldescriptor, making the \textit{mmkg:filePath} data property redundant, as it is inherited from \modaldescriptor. 
Additionally, \textit{dc:format} is moved from \textit{Scene} to the \modality class, alongside \textit{dc:sizeOrDuration}, which is further broken down into width, height, and duration based on the type of modality. 
The \textit{Scene} encompasses the role of the \mmentity, and, consequently, this class inherits the \hasmodaldescriptor property, which connects the Scene with its associated resource(s).


\subsubsection{Polifonia Ontology Network (PON)}
\label{subsubsec:Polifonia}
In the domain of music, several ontologies have been proposed to annotate both symbolic and audio data, aiming to create comprehensive Music Knowledge Graphs. 
However, existing models suffer from a lack of interoperability and fail to adequately capture the historical and cultural contexts surrounding music creation. 
This limitation might also propagate bias toward recent trends and cultural preferences in downstream applications.
As a response to these potential issues, \cite{Berardinis2023} introduced the Polifonia Ontology Network (PON), offering a modular framework of music ontologies to address both cultural heritage preservation and broader queries within the music domain. 
Currently, PON is composed of 15 modules, each addressing various aspects of the music domain. 
Notably, the CoMeta module, an extension of Music Meta \cite{de2023music}, focuses on describing the metadata of music collections, corpora, containers, or music datasets, with a particular emphasis on multi-modal concepts.

\begin{figure}[!t]
\centering
    \includegraphics[scale=0.44]{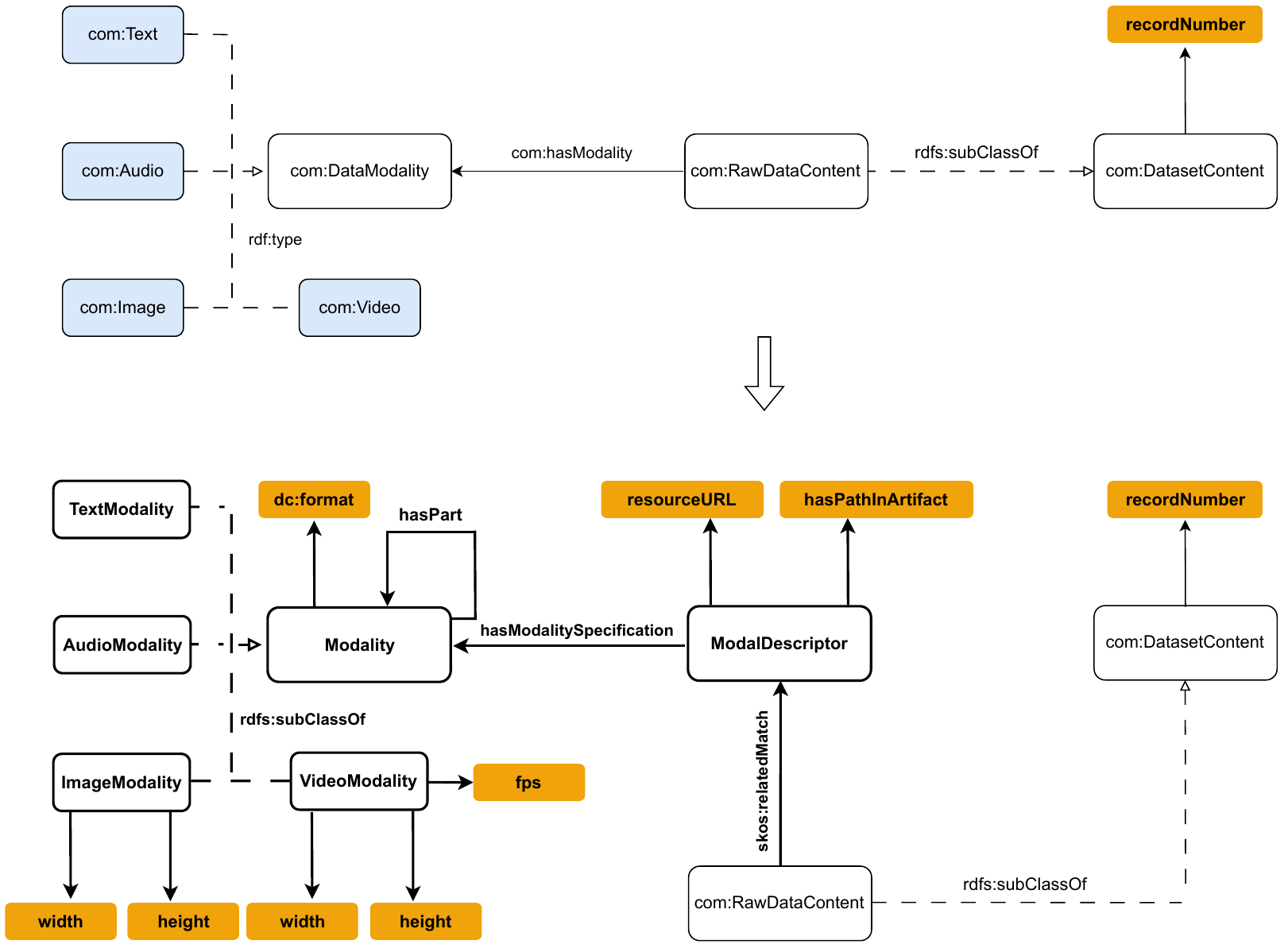}
    \caption{Alignment of the multi-modal pattern with the CoMeta ontology module in PON \cite{Berardinis2023}.}
    \label{fig:alignment_with_polifonia}
\end{figure}

The upper part of Figure~\ref{fig:alignment_with_polifonia} illustrates an excerpt of the CoMeta ontology, depicting the relationship between the primary concepts involving multi-modality: \textit{com:DatasetContent}, \textit{com:RawDataDocument} and \textit{com:Modality}. 
The \textit{DatasetContent} describes the content of a dataset from a summative perspective (e.g. the audio content of a music collection, the audio features it provides, etc.) and its production process (provenance). 
The \textit{RawDataContent} class, a subclass of \textit{DatasetContent}, is characterised as a partition of the dataset containing raw data of a specific modality, whether structured (e.g., tabular data) or unstructured (e.g., audio files).
For example, a dataset partition containing images of music albums could be described as \textit{RawDataContent}.
Finally, the \textit{RawDataContent} class is linked, via the \textit{hasModality} object property, to the \textit{DataModality} class, which describes the modality of a dataset partition, such as audio, video, or image. 
The connection between \textit{RawDataContent} and \textit{DataModality} strongly aligns with that of \modaldescriptor and \modality in the Multi-modal pattern in Figure~\ref{fig:conceptualization_mm}. The main difference is that instances of single modalities are not as specific as those in the ODP.

The alignment of CoMeta within the ODP is illustrated in Figure~\ref{fig:alignment_with_polifonia} (\textit{bottom}).
As can be seen from the figure, \textit{RawDataContent} class is now connected to the \modaldescriptor by the object property \textit{skos:relatedMatch}\footnote{\href{https://www.w3.org/2009/08/skos-reference/skos.html\#relatedMatch}{https://www.w3.org/2009/08/skos-reference/skos.html\#relatedMatch}}, which is used to state an associative mapping link between the two concepts, and the \hasmodality relation which originates from the \modaldescriptor class. 
As a consequence, single modalities are no longer just instances but conceptual entities serving the objectives outlined in Section~\ref{sec:04-pattern}.

As we find no \mmentity in CoMeta, this example demonstrates the partial reuse of the Multi-modal pattern when the focus is on the modality specifications -- to describe the various realisations in a dataset, rather than on the information objects.


\section{Conclusion and Future Works}
\label{sec:06-conclusion}


This article proposes a novel multi-modal ODP that facilitates the integration of different modalities to define multi-modal entities. Our design is based on a comprehensive analysis of how multi-modal knowledge has been represented in the context of knowledge representation (Section \ref{sec:02-background}). Specifically, we focused our analysis on the construction and use of MMKGs, and we have identified three main limitations of current approaches (Section~\ref{sec:03-motivation}) that helped us scope the requirements that a multi-modal ODP should satisfy (Section~\ref{sec:04-pattern}).

The proposed pattern aims to foster interoperability among diverse systems, laying the foundation for more cohesive and integrated (multi-modal) knowledge graphs. 
Through the application of our pattern to existing knowledge graphs and its alignments with other ontologies found in the literature (Section~\ref{sec:05-application}), we have shown both its usability and generalisability.

It is worth mentioning that during the alignment of external ontologies with our pattern, we encountered difficulties in accessing human-readable documentation detailing the concepts used within these ontologies. 
As in many types of ontological reuse, the accurate alignment of these resources depends on the availability of detailed documentation that describes the intent of the ontology developers and examples of how the ontology can be used to define certain notions~\cite{tudorache2020ontology,alharbi2021characterising}.
This highlights the importance of both human-readable documentation and concrete examples in facilitating a clear understanding of an ontology's concepts, thereby mitigating the risk of design errors resulting from misinterpreting concepts. Therefore, the alignments shown in Section~\ref{sec:05-application}, represent our best effort given the available resources. 
However, we do not exclude that with more detailed documentation and examples for the aforementioned ontologies, these alignments could be potentially refined further.

Moving forward, there are several directions for further refinement and exploration. 
First, extending the validation process described in Section~\ref{sec:05-application} to include a wider range of external ontologies, will enhance the robustness and applicability of our pattern. Second, refining the multi-modal ODP by introducing additional modalities and their associated data properties will broaden its applicability across various domains. Lastly, modelling the relationships between the different modalities, such as text-to-audio conversion, as well as including user preferences and constraints over modalities, will enhance the adaptability and the user-centric nature of multi-modal knowledge representation systems. 

\clearpage 
%
%
%

\bibliographystyle{splncs04}
\bibliography{bibliography}

\end{document}